\documentclass[sigconf]{acmart}

\AtBeginDocument{%
  \providecommand\BibTeX{{%
    \normalfont B\kern-0.5em{\scshape i\kern-0.25em b}\kern-0.8em\TeX}}}

\setcopyright{none}
\copyrightyear{2022}
\acmYear{2022}

\renewcommand\footnotetextcopyrightpermission[1]{}

\usepackage{algorithmic}
\usepackage{graphicx}
\usepackage{textcomp}
\usepackage{xcolor,colortbl}

\usepackage{circuitikz}
\usetikzlibrary{shapes.geometric, patterns, arrows.meta, shapes.arrows, fit, shadows.blur}
\usepackage{pgfplots,pgfplotstable}
\usepackage{subfig}
\usepackage{threeparttable}
\usepackage{multirow,hhline}
\usepackage{xcolor, colortbl}
\usepackage{todonotes}
\usepackage{setspace}

\usepackage{collcell}
\usepackage{enumitem}
\usepackage{tikz}
\usetikzlibrary{matrix}
\usepackage{multirow}

\definecolor{babyblue}{rgb}{0.54,0.81,0.94}
\definecolor{mypink}{rgb}{0.858, 0.188, 0.478}

\usepackage{PieChart}
\usetikzlibrary{decorations.text}

\usepackage{enumitem}

\usepackage{dblfloatfix}

\pgfplotsset{compat=1.18}

\begin{document}

%%
%% The "title" command has an optional parameter,
%% allowing the author to define a "short title" to be used in page headers.
\title{TsetlinWiSARD: On-Chip Training of Weightless Neural Networks using Tsetlin Automata on FPGAs}

\makeatletter
\def\ps@plain{%
  \def\@oddhead{\small PREPRINT - Accepted at the 63rd ACM/IEEE Design Automation Conference (DAC 2026)\hfill\thepage}
  \def\@evenhead{\small PREPRINT - Accepted at the 63rd ACM/IEEE Design Automation Conference (DAC 2026)\hfill\thepage}
  \def\@oddfoot{}
  \def\@evenfoot{}
}
\pagestyle{plain}
\makeatother

%%
%% The "author" command and its associated commands are used to define
%% the authors and their affiliations.
%% Of note is the shared affiliation of the first two authors, and the
%% "authornote" and "authornotemark" commands
%% used to denote shared contribution to the research.
\author{
Shengyu Duan$^{*}$, Marcos L. L. Sartori$^{*}$, Rishad Shafik$^{*}$, Alex Yakovlev$^{*,\dagger}$\\
$^{*}$Microsystems Research Group, Newcastle University \ \ $^{\dagger}$ Literal Labs, Newcastle upon Tyne, UK\\
\{shengyu.duan,  marcos.sartori,  rishad.shafik,  alex.yakovlev\}@newcastle.ac.uk
}

\begin{abstract}
	Increasing demands for adaptability, privacy, and security at the edge have persistently pushed the frontiers for a new generation of machine learning (ML) algorithms with training and inference capabilities on-chip. Weightless Neural Network (WNN) is such an algorithm that is principled on lookup table based simple neuron structures. As a result, it offers architectural benefits, such as low-latency, low-complexity inference, compared to deep neural networks that depend heavily on multiply–accumulate operations. However, 
    traditional WNNs rely on memorization-based one-shot training, which either leads to overfitting and reduced accuracy or requires tedious post-training adjustments, limiting their effectiveness for efficient on-chip training.
    %current WNN training methods use approximations that either reduce accuracy or require tedious post-training adjustments, limiting their suitability for efficient on-chip training.
	
	In this work, we propose TsetlinWiSARD, a training approach for WNNs that leverages Tsetlin Automata (TAs) to enable probabilistic, feedback-driven learning.
    It overcomes the overfitting of WiSARD’s one-shot training with iterative optimization, while maintaining simple, continuous binary feedback for efficient on-chip training.
    Central to our approach is a field programmable gate array (FPGA)-based training architecture that delivers state-of-the-art accuracy while significantly improving hardware efficiency. Our approach provides over 1000$\times$ faster training when compared with the traditional WiSARD implementation of WNNs. Further, we demonstrate 22\% reduced resource usage, 93.3\% lower latency, and 64.2\% lower power consumption compared to FPGA-based training accelerators implementing other ML algorithms.
\end{abstract}

%%
%% Keywords. The author(s) should pick words that accurately describe
%% the work being presented. Separate the keywords with commas.
\keywords{Machine learning, Weightless neural networks, FPGA, Tsetlin automata, On-chip training}

%%
%% This command processes the author and affiliation and title
%% information and builds the first part of the formatted document.
\maketitle
\thispagestyle{plain}

\section{Introduction} \label{sec:intro}
Demands for fast processing, privacy, and security have driven machine learning (ML) from cloud-centric architectures toward edge computing, where resources are highly constrained. These challenges have been extensively studied for ML inference, in which models are pre-trained—often with algorithm-level compression—and then deployed on edge \cite{li2023comprehensive}. However, many real-world services, such as sensor-driven applications, require real-time model adaptation to cope with dynamically changing environments \cite{papst2024sensor}. In such settings, deploying inference alone becomes inefficient due to the costly and continuous communication between cloud-based training and edge-based inference nodes \cite{tang2022ef}. This motivates the need for on-chip training, enabling edge devices to learn continuously and locally from new data, but it also greatly intensifies the constraints of edge hardware; for example, training deep learning models can consume over 200$\times$ more energy than inference \cite{aquino2025towards}.

Many ML models tend to be domain-specific, with data, architectures, and optimization strategies customized for each application. This often necessitates redesign or reconfiguration of the hardware implementation. FPGAs therefore offer a key advantage: their rapid, flexible reconfigurability enables much faster development and adaptation than time-consuming ASIC design flow, making them well-suited for evolving ML deployments.
%especially for runtime adaptation or on-chip training enabled by reconfiguration.

Relatively few studies have investigated ML on-chip training on FPGAs, with most efforts centered on deep neural networks (DNNs) \cite{tang2022ef, zhao2016f, luo2020towards}. Yet the intrinsic algorithmic demands of DNN training, including large amounts of floating-point multiplication and gradient computation, are difficult to support efficiently on FPGAs, whose compute resources are fundamentally designed around fixed-point arithmetic \cite{mao2025dynamic}.

%Weightless Neural Networks (WNNs) offer an efficient alternative to DNNs, operating without multiplication and relying instead on memory-based pattern recognition \cite{miranda2022logicwisard}. \todo{RS: For those who do not know WNN well, distinguish WNNs from traditional NNs}
Weightless Neural Networks (WNNs) offer a promising alternative to DNNs. Unlike DNNs, WNNs do not rely on multiplication but instead perform memory-based pattern recognition \cite{miranda2022logicwisard}.
A typical WNN is composed of neurons implemented as binary lookup tables (LUTs)—also referred to as RAM nodes—that recognize binary input patterns through direct LUT indexing.
%The regular LUT-based structure of WNNs maps efficiently onto FPGAs, whose architecture is fundamentally built on LUTs.\todo{RS: Why is a study of WNN worth it? What benefits does it provide?}
This key difference makes WNNs a natural fit for efficient mapping onto FPGA hardware, as FPGAs are inherently built around LUTs. This structural alignment leads to reduced computational complexity and lower resource consumption compared to DNNs.

%One of the most extensively studied WNN models is the WiSARD (Wilkie, Stonham, and Aleksander Recognition Device), exhibiting key characteristics such as one-shot training, a separate discriminator per class, and a shallow, one-layer architecture \cite{aleksander1984wisard}. These features enable fast training, low-latency and low-power inference, and cost-efficient implementation.
%However, the pattern memorization-based training of WiSARD and many other WNN models often leads to overfitting and thus lower accuracy compared to modern DNNs, particularly on complex tasks \cite{susskind2022weightless}.
%However, the simplicity of WiSARD
%\todo{simplcity does not cause lower accuracy -- perhaps pruning some computational units may do -- please check.}
%and many other WNN models often leads to lower accuracy compared to modern DNNs, particularly on complex tasks \cite{susskind2022weightless}.
%As a result, they are better suited for applications where constraints such as cost or latency are prioritized over absolute accuracy, such as those in edge computing and embedded systems \cite{susskind2022weightless, miranda2022logicwisard, ferreira2019feasible}.
%As a result, they remain primarily suited for applications where constraints such as cost or latency are prioritized over absolute accuracy \cite{susskind2022weightless, miranda2022logicwisard, ferreira2019feasible}.

One of the most extensively studied WNN models is the WiSARD (Wilkie, Stonham, and Aleksander Recognition Device), which features memorization-based training, a separate discriminator per class, and a one-layer architecture \cite{aleksander1984wisard}.
%\todo{Say what are the key challenges this model has and then motivate your inference plus training infrastructure.}
We present the inference structure of WiSARD in Figure~\ref{fig:tm_overview}. These properties potentially enable on-chip training, low-latency and low-power inference, and cost-efficient implementation \cite{ferreira2019feasible}.
However, WiSARD faces several key challenges.
Firstly, its reliance on pattern memorization often leads to overfitting, reducing accuracy on complex tasks \cite{susskind2022weightless}.
%However, the pattern-memorization nature of the traditional WiSARD often leads to overfitting and therefore lower accuracy on complex tasks \cite{susskind2022weightless}. 
Secondly, the only more advanced training method currently available—B-bleaching—relies on interactive post-training parameter selection, which cannot be integrated into on-chip training \cite{bleaching}.
These challenges highlight the need for an inference and training infrastructure that can overcome both the overfitting issue and the inability to support on-chip adaptation.
A more detailed explanation of existing WiSARD training methods is provided in Section~\ref{sec:wnn}.
%To date, no training approach for WiSARD achieves state-of-the-art accuracy while remaining compatible with on-chip training.

\definecolor{mypink}{rgb}{0.858, 0.188, 0.478}

\newcommand{\overbar}[1]{\mkern 2.5mu\overline{\mkern-2.5mu#1\mkern-2.5mu}\mkern 2.5mu}

\newcommand*{\MinNumberG}{0}%
\newcommand*{\MaxNumberG}{16}%

\newcommand{\ApplyGradientG}[1]{%
	\pgfmathsetmacro{\PercentColor}{100.0*(#1-\MinNumberG)/(\MaxNumberG-\MinNumberG)}
	\hspace{-0.33em}\colorbox{white!\PercentColor!black}{}
}

\newcolumntype{G}{>{\collectcell\ApplyGradientG}c<{\endcollectcell}}

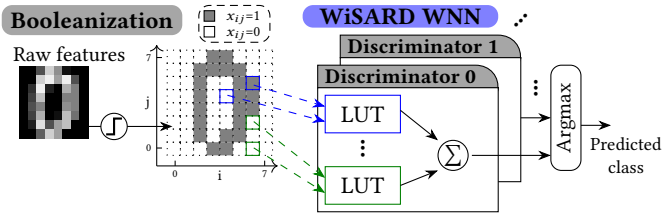
\begin{figure}[!htb]
%\vspace{-0.5cm}

\setlength{\fboxsep}{3mm} % box size
\setlength{\tabcolsep}{0pt}
\centering
	\begin{tikzpicture}[every node/.style={scale=1}, scale=1]
	
\tikzset{square matrix/.style={
		matrix of nodes,
		column sep=-\pgflinewidth, row sep=-\pgflinewidth,
		nodes={draw,
			dotted,
			minimum height=#1,
			anchor=center,
			text width=#1,
			align=center,
			inner sep=0pt
		},
	},
	square matrix/.default=2.5mm
}

% Booleanization
\begin{scope}[xshift=0cm,yshift=-2.9cm,every node/.style={scale=1}, scale=1]
\node[] at (0.2,0.65) {\small{Raw features}};
\draw[draw=none, fill=gray!70!white, rounded corners] (-0.7,0.9) rectangle ++(2.4,0.4) node[pos=.5] () {\textbf{Booleanization}};
		
	\begin{scope}[xshift=0cm,yshift=0cm,every node/.style={scale=0.2}, scale=0.2]
		\node[] at (0,0) {
		\begin{tabular}{*{8}{G}}
0&0&5&13&9&1&0&0\\
0&0&13&15&10&15&5&0\\
0&3&15&2&0&11&8&0\\
0&4&12&0&0&8&8&0\\
0&5&8&0&0&9&8&0\\
0&4&11&0&1&12&7&0\\
0&2&14&5&10&12&0&0\\
0&0&6&13&10&0&0&0\\
		\end{tabular}
	};
\end{scope}
\end{scope}

\begin{scope}[xshift=2.2cm]
\begin{scope}[xshift=0cm,yshift=-2.9cm,every node/.style={scale=0.7}, scale=0.7]
\matrix[square matrix]
{
||&||&||&|[fill=gray]|&|[fill=gray]|&||&||&||\\
||&||&|[fill=gray]|&|[fill=gray]|&|[fill=gray]|&|[fill=gray]|&||&||\\
||&||&|[fill=gray]|&||&||&|[fill=gray]|&|[fill=gray]|&||\\
||&||&|[fill=gray]|&||&||&|[fill=gray]|&|[fill=gray]|&||\\
||&||&|[fill=gray]|&||&||&|[fill=gray]|&|[fill=gray]|&||\\
||&||&|[fill=gray]|&||&||&|[fill=gray]|&||&||\\
||&||&|[fill=gray]|&||&|[fill=gray]|&|[fill=gray]|&||&||\\
||&||&||&|[fill=gray]|&|[fill=gray]|&||&||&||\\
};

\draw[thin, blue] (0.5,0.25) rectangle ++(0.25,0.25) node[] () {};
\draw[thin, blue] (0,0) rectangle ++(0.25,0.25) node[] () {};

\draw[thin, green!50!black] (0.5,-0.5) rectangle ++(0.25,0.25) node[] () {};
\draw[thin, green!50!black] (0.5,-1) rectangle ++(0.25,0.25) node[] () {};

\draw[->] (-1.2,-1.2) -- (1.1,-1.2);
\draw[->] (-1.2,-1.2) -- (-1.2,1.1);
\draw[] (-0.85,-1.2) -- (-0.85,-1.1)
(-1.2,-0.86) -- (-1.1,-0.86)
(0.85,-1.2) -- (0.85,-1.1)
(-1.2,0.86) -- (-1.1,0.86)
;
\node[] at (-0.85,-1.35) {\scriptsize{0}};
\node[] at (0.85,-1.35) {\scriptsize{7}};
\node[] at (-1.35,-0.86) {\scriptsize{0}};
\node[] at (-1.35,0.86) {\scriptsize{7}};
\node[] at (0,-1.37) {\small{i}};
\node[] at (-1.37,0) {\small{j}};

\begin{scope}[xshift=1cm]
\draw[] (-1.3,1.25) rectangle ++(0.2,0.2) node[] () {};
\node[] at (-0.55,1.32) {\small{$x_{ij}$=0}};
\draw[fill=gray] (-1.3,1.55) rectangle ++(0.2,0.2) node[] () {};
\node[] at (-0.55,1.62) {\small{$x_{ij}$=1}};

\draw[rounded corners,densely dashed] (-1.4,1.15) rectangle ++(1.35,0.7) node[] () {};
\end{scope}

\begin{scope}[xshift=-2.8cm, yshift=-2.2cm]
\draw[-Stealth] (0.2,1.75) --++ (1.7,0);
\node[draw,circle,minimum size=0.5cm,inner sep=0pt,fill=white] at (0.79,1.75) {};
\draw[thick] (0.94,1.9) --++ (-0.15,0) --++ (0,-0.3) --++ (-0.15,0);
\end{scope}
\end{scope}

% WNN
\begin{scope}[xshift=3.6cm,yshift=-2.6cm]

\draw[fill=gray!70!white] 
(0.45,0.6) -- (0.4,0.3) -- (-2,0.3)   -- (-2,0.6) [rounded corners=10pt] -- cycle;
\draw[fill=white] (-2,-1.35) rectangle ++ (2.4,1.65) node[] {};
\node[] at (-0.9,0.45) {\textbf{\small{Discriminator 1}}};

\end{scope}

\begin{scope}[xshift=3.3cm,yshift=-3cm]

\draw[fill=gray!70!white] 
(0.45,0.6) -- (0.4,0.3) -- (-2,0.3)   -- (-2,0.6) [rounded corners=10pt] -- cycle;
\draw[fill=white] (-2,-1.35) rectangle ++ (2.4,1.65) node[] {};
\node[] at (-0.9,0.45) {\textbf{\small{Discriminator 0}}};

\draw
(-1.4, -0.05) node[] () {LUT}
;
\draw[blue] (-1.9,-0.3) rectangle ++ (1,0.5) node[] {};

\draw[blue, dashed, -Stealth] (-2.85,0.335) -- (-1.9,0.05);
\draw[blue, dashed, -Stealth] (-3.2,0.2) -- (-1.9,-0.15);

\draw
(-1.4, -1) node[] () {LUT}
;
\draw[green!50!black] (-1.9,-1.25) rectangle ++ (1,0.5) node[] {};

\draw[green!50!black, dashed, -Stealth] (-2.85, -0.15) -- (-1.9,-0.9);
\draw[green!50!black, dashed, -Stealth] (-2.85, -0.5) -- (-1.9,-1.1); 

\node[rotate=90] at (-1.4,-0.5) {\large{\textbf{...}}};

\draw[-Stealth] (-0.9,0) -- (-0.4,-0.4);
\draw[-Stealth] (-0.9,-1) -- (-0.4,-0.8);

\node[draw,circle,minimum size=0.4cm,inner sep=0pt] at (-0.2,-0.6) {$\sum$};
\draw[-Stealth] (0.05,-0.6) -- (1.1,-0.6);
\draw[-Stealth] (0.7,-0.1) -- (1.1,-0.1);
\node[rotate=90] at (0.9,0.3) {\large{\textbf{...}}};

\draw[rounded corners] (1.1,-0.8) rectangle ++(0.4,1.4) node[pos=.5, rotate=90] () {\small{Argmax}};

\draw[-Stealth] (1.5,-0.2) -- (1.9,-0.2);

\node[] at (2.1, -0.45) {\footnotesize{Predicted}};
\node[] at (2.1, -0.75) {\footnotesize{class}};

\node[rotate=45] at (0.7,1.2) {\large{\textbf{...}}};

\draw[draw=none, fill=blue!50!white, rounded corners] (-2.2,1.1) rectangle ++(2.6,0.3) node[pos=.5] () {\textbf{WiSARD WNN}};

\end{scope}
\end{scope}
	
	\end{tikzpicture}
\caption{Inference structures of WiSARD. WiSARD operates on Boolean input features, typically derived by thresholding raw data, and performs classification through a voting mechanism. It needs one discriminator per class. The voting units are LUTs, each receiving a fixed feature subset randomly mapped from the full feature set during training initialization and capable of representing any Boolean function over that subset. All discriminators use the same feature shuffle.
}
\label{fig:tm_overview}
\vspace{-0.6cm}
\end{figure}

To tackle the above challenges in parallel, we propose TsetlinWiSARD, which introduces
\begin{enumerate}[itemsep=0pt, topsep=0pt, partopsep=0pt, parsep=0pt]
\item a new training approach for WiSARD based on Tsetlin Automata (TAs) to enhance learning capability and significantly improve accuracy, and
\item on-chip training architectures that leverage the hardware-efficient design of TsetlinWiSARD, enabling 1000$\times$ faster, processor-free training on FPGA.
\end{enumerate}
%\todo{remove white space around the *enumerate*}

Specifically, a TA is a finite-state learning automaton that selects optimal actions in stochastic environments based on binary feedback \cite{tsetlin1961behaviour}. Its internal states are divided into two zones, each corresponding to a possible action, and state transitions reinforce actions that receive positive feedback. 

On the algorithm front, TA-based adaptive updates enable iterative, optimization-driven learning, allowing TAs to move WiSARD beyond conventional memorization and toward improved accuracy–cost trade-offs. Across six benchmark datasets, we show TsetlinWiSARD consistently achieves significantly higher accuracy than existing WiSARD training methods when models of comparable size are used. An additional empirical study investigates the impact of TsetlinWiSARD’s structural and learning parameters on classification performance, yielding guidelines for the principled selection of training parameters.

On the hardware front, the proposed on-chip training architecture maps WiSARD’s logical LUTs onto the physical LUTs of FPGAs and implements TAs by compacting 2$^n$ automata into a small number of $n$-input LUTs. Training is performed with continous, binary feedback using simple hardware operations such as logical comparison, increment, and decrement. Implemented on a Xilinx XC7Z020 FPGA, the architecture achieves over 1000$\times$ faster training compared to the only existing WiSARD training accelerator, and represents the first processor-free WiSARD accelerator that supports training. In comparison to FPGA-based training accelerators for other ML alorithsm-Tsetlin Machine (TM) and Convolutional Neural Networks (CNNs), TsetlinWiSARD uses over 22.0\% fewer hardware resources, reduces training latency by at least 93.3\%, and lowers power consumption by at least 64.2\%, makes TsetlinWiSARD ideal for edge ML applications.

%The hardware-friendly design of TsetlinWiSARD makes it well-suited for on-chip training, particularly on FPGAs. We propose an on-chip training architecture that maps WiSARD's logical LUTs directly onto the physical LUTs of the FPGA. This design efficiently implements TAs by compacting 2$^n$ TAs into a small number of $n$-input LUTs. Implemented on a Xilinx XC7Z020 FPGA, the architecture achieves over 1000$\times$ faster training compared to the only existing WiSARD training accelerator, and represents the first processor-free WiSARD accelerator that supports training. In comparison to FPGA-based training accelerators for TM and Convolutional Neural Networks (CNNs), TsetlinWiSARD uses over 22.0\% fewer hardware resources, reduces training latency by at least 93.3\%, and lowers power consumption by at least 64.2\%. TsetlinWiSARD’s ability to perform fast, low-cost on-chip learning makes it ideal for edge ML applications.

%The key \textit{contributions} are as follows:
%\begin{itemize}
%	\item A TA-based training method that significantly improves WiSARD accuracy.
%	\item Extensive empirical analysis uncovering key factors for optimizing TsetlinWiSARD training.
%	\item A compact on-chip architecture enabling 1000$\times$ faster, processor-free WiSARD training on FPGA.
%\end{itemize}

Finally, it is worth noting that the use of TAs as learning elements was first demonstrated in the TM, a recently proposed machine learning model~\cite{granmo2018tsetlin}. However, the learning mechanism of TsetlinWiSARD fundamentally differs from TMs: In TMs, TAs decide whether a feature should be included in a conjunctive clause~\cite{granmo2018tsetlin}. In contrast, WiSARD learns the values stored in its LUTs, making the underlying learning objective inherently distinct. We show that TsetlinWiSARD utilizes TAs more efficiently: it achieves accuracy competitive with or superior to that of TMs while requiring more than one-third fewer TAs on most benchmarks.

%The remainder of this paper is organized as follows. Section~\ref{sec:wnn} reviews the state-of-the-art in WNNs. Section~\ref{sec:twnn} presents the TsetlinWiSARD training mechanism and its hardware architecture. Sections~\ref{sec:off-chip} and~\ref{sec:on-chip} report the off-chip and on-chip experimental results, respectively, along with comparisons to existing ML algorithms and hardware accelerators. Finally, the paper is concluded in Section~\ref{sec:con}.

This paper is organized as follows: Section~\ref{sec:wnn} reviews WNNs, Section~\ref{sec:twnn} presents the TsetlinWiSARD training mechanism and architecture, Sections~\ref{sec:off-chip} and~\ref{sec:on-chip} report off-chip and on-chip results, respectively, with comparisons to existing ML algorithms and hardware accelerators. Finally, Section~\ref{sec:con} concludes the paper.

\vspace{-0.2cm}
\section{Weightless Neural Networks} \label{sec:wnn} 
%\todo{Include a short para to introduce to contextualise what you are presenting in this section.}\textbf{Training}: 
%\todo{First say what purpose training serves in ML for non-ML reviewers.}
\textbf{Training}, serves to adjust model parameters so it can recognize patterns on unseen data.
WiSARD performs one-shot training, where all LUT entries are initialized to zero, and those corresponding to the training sample in the correct-class discriminator are set to one. This immediately memorizes all seen patterns without iterative updates \cite{aleksander1984wisard}.
%\todo{Insert reference.}
However, with large training sets, most LUT entries become one, causing saturation—all discriminators produce similar vote counts, leading to frequent classification ties.

To address this, the B-bleaching algorithm replaces binary LUTs with integer counters that record pattern frequencies during training \cite{bleaching}. A post-training statistical analysis determines a threshold to convert these counts back to binary for inference, restoring the original WiSARD structure. 
However, B-bleaching still remains a pattern memorization method without an iterative refinement process.
In addition, the need for post-training thresholding renders B-bleaching impractical for real-time, on-chip learning.

%\textbf{Inference}: 
%\todo{First say what inference means to a non-ML reviewer.}
\textbf{Inference}, is to use trained model to make predictions on new data.
Several WiSARD-like architectures have been proposed to improve inference efficiency, primarily through post-training model compression. These approaches exploit the high sparsity of LUTs, where most entries remain unused during training. To reduce memory usage, LUTs are often replaced with hash tables or Bloom filters, which store exact or approximate membership of activated patterns ($i.e.$, updated entries) \cite{ferreira2019feasible, susskind2022weightless, santiago2018memory, susskind2023uleen}. Other compression methods include merging LUTs across discriminators and converting LUTs into Boolean functions \cite{miranda2022logicwisard}. Despite these improvements in inference efficiency, the underlying training process in these architectures still relies on standard WiSARD or B-bleaching.

\textbf{Recent advancements} in WNNs have introduced gradient-based training approaches, such as the straight-through estimator for single-layer WNNs \cite{susskind2023uleen} and finite difference for multi-layer architectures \cite{bacellar2024differentiable}, achieving significantly improved accuracy. 
%\todo{Perhaps highlight the key approaches and limitations first rather than saying these are not the focus.}
%However, these works are not the focus of this paper, as we target fully discrete learning, where training cost and hardware simplicity are the primary concerns. Our approach emphasizes lightweight, efficient training over accuracy gains achieved through gradient approximations.
However, this also comes at the cost of high computational overhead from gradient approximations and increased architectural complexity, making on-chip training and hardware implementations significantly more resource-intensive and less suitable for real-time or edge applications.
\vspace{-0.2cm}
\section{TsetlinWiSARD and On-Chip Training} \label{sec:twnn}

\subsection{Training Mechanism} \label{sec:training}
%\todo{Since this is your main contribution -- make that clear. Also, connect with your previous arguments in the previous section.}
Unlike previous works, we enhance WiSARD’s learning capability by integrating TAs for lightweight, fully discrete learning—without adding structural complexity for inference:
The inference architecture of TsetlinWiSARD remains identical to that of standard WiSARD, as illustrated in Figure~\ref{fig:tm_overview}. The key difference lies in the training scheme: in TsetlinWiSARD, TAs act as learning elements integrated into each LUT to determine the values of all entries. These TAs are only involved during training and are not required for inference. As a result, existing inference optimization techniques for WiSARD—such as those described in Section \ref{sec:wnn}—remain fully applicable to models trained using TsetlinWiSARD.

The organization of TAs is given in Figure~\ref{fig:LUT}. For each LUT, one TA is assigned to every entry, with its internal state determining the output value: states below/above the midpoint represent output 0/1. An $n$-input LUT requires 2$^n$ TAs to cover all addresses.

\definecolor{new_red}{rgb}{0.95,0.47,0.47}
\definecolor{new_green}{rgb}{0.85,0.98,0.77}

\definecolor{dark_red}{rgb}{0.78,0.28,0.34}
\definecolor{dark_green}{rgb}{0.09,0.30,0.28}

\tikzset{
	sectors/.style n args={6}{minimum width=#1, minimum height=#2, 
		text depth=0.25ex, outer sep=0pt,
		append after command={\pgfextra{\let\LN\tikzlastnode
				\draw[fill=#3]  (\LN.west)  -| (\LN.north)
				{[rounded corners=3mm] -- (\LN.north west)}
				{[rounded corners=3mm] -- cycle}
				;
				
				\draw[fill=#5]  (\LN.east)  -| (\LN.north)
				{[rounded corners=3mm] -- (\LN.north east)}
				[rounded corners=3mm] -- cycle;
				;
	} } }
}

\begin{figure}[!htb]
\vspace{-0.4cm}
\centering
\begin{tikzpicture}
% LUT
\node[] at (0.6,0.6) {Address};
\draw[] (0,0) rectangle ++(1.2,0.4) node[pos=.5] () {00...0};
\node[] at (1.8,0.6) {Output};
\draw[] (1.2,0) rectangle ++(1.2,0.4) node[pos=.5] () {1};
\node[] at (1.2,1.1) {\textbf{LUT}};
\node[] at (5.2,0.9) {\textbf{Tsetlin Automata}};

\draw[] (0,0) rectangle ++(1.2,-1.2) node[pos=.5,rotate=90] {...};
\draw[] (1.2,0) rectangle ++(1.2,-1.2) node[pos=.5,rotate=90] {...};
\node[rotate=90] at (5.15,-0.6) {...};
\node[rotate=90] at (-0.7, -0.6) {\small{$2^n$}};
\node[rotate=90] at (-0.3, -0.6) {\small{entries/TAs}};
\draw[gray] (-0.5,0.1) -- (-0.5,0.4)
(-0.6,0.4) -- (-0.4,0.4)
(-0.5,-1.3) -- (-0.5,-1.6)
(-0.6,-1.6) -- (-0.4,-1.6); 

\draw[] (0,-1.2) rectangle ++(1.2,-0.4) node[pos=.5] () {11...1};
\draw[] (1.2,-1.2) rectangle ++(1.2,-0.4) node[pos=.5] () {0};

\node[] at (0.6,-1.8) {\small{$n$-bit}};
\draw[gray] (0.2,-1.8) -- (0,-1.8)
(0,-1.9) -- (0,-1.7)
(1,-1.8) -- (1.2,-1.8)
(1.2,-1.9) -- (1.2,-1.7); 

% TA 0
\begin{scope}[xshift=3.5cm,yshift=0.2cm, scale=0.6, every node/.style={scale=0.6}]
\node at (2.75,-0.7) [sectors= {72mm}{28mm}
{new_red}{}
{new_green}{}] {};

\draw[thick,fill=white] (0,0) circle [radius=0.3] node (s1) {1};
\draw[thick, arrows={-Stealth[reversed,reversed]},dashed] (-0.3,-0.2) to [out=230,in=160,looseness=4] (-0.3,0.2);
\draw[thick,fill=white] (1,0) circle [radius=0.3] node (s2) {2};
\draw[thick, arrows={-Stealth[reversed,reversed]}] (0,0.3) to [bend left=45] (1,0.3);
\draw[thick, arrows={Stealth[reversed,reversed]-},dashed] (0,-0.3) to [bend right=45] (1,-0.3);
\draw[thick,fill=white] (2.2,0) circle [radius=0.3] node (sn) {N};
\node[] at (1.6,0) {...};
\draw[thick, arrows={-Stealth[reversed,reversed]}] (1,0.3) to [bend left=35] (2.2,0.3);
\draw[thick, arrows={Stealth[reversed,reversed]-},dashed] (1,-0.3) to [bend right=35] (2.2,-0.3);

\draw[thick, arrows={-Stealth[reversed,reversed]}] (2.2,0.3) to [bend left=35] (3.3,0.3);
\draw[thick, arrows={Stealth[reversed,reversed]-},dashed] (2.2,-0.3) to [bend right=35] (3.3,-0.3);

\draw[thick,fill=white] (3.3,0) circle [radius=0.3] node (sn1) {\footnotesize{N+1}};
\draw[thick, arrows={-Stealth[reversed,reversed]}] (3.3,0.3) to [bend left=45] (4.3,0.3);
\draw[thick, arrows={Stealth[reversed,reversed]-},dashed] (3.3,-0.3) to [bend right=45] (4.3,-0.3);
\draw[thick,fill=white] (4.3,0) circle [radius=0.3] node (sn2) {\footnotesize{N+2}};
\node[] at (4.9,0) {...};
\draw[thick, arrows={-Stealth[reversed,reversed]}] (4.3,0.3) to [bend left=35] (5.5,0.3);
\draw[thick, arrows={Stealth[reversed,reversed]-},dashed] (4.3,-0.3) to [bend right=35] (5.5,-0.3);
\draw[thick,fill=white] (5.5,0) circle [radius=0.3] node (s2n) {};
\draw[draw=none,fill=black!25!white] (5.5,0) circle [radius=0.24] node () {2N};
\draw[thick, arrows={-Stealth[reversed,reversed]}] (5.8,0.2) to [out=30,in=330,looseness=4] (5.8,-0.2);
\end{scope}

% last TA
\begin{scope}[xshift=3.5cm,yshift=-1.4cm, scale=0.6, every node/.style={scale=0.6}]
\node at (2.75,-0.7) [sectors= {72mm}{28mm}
{new_red}{}
{new_green}{}] {};

\draw[thick,fill=white] (0,0) circle [radius=0.3] node (s1) {1};
\draw[thick, arrows={-Stealth[reversed,reversed]},dashed] (-0.3,-0.2) to [out=230,in=160,looseness=4] (-0.3,0.2);
\draw[thick,fill=white] (1,0) circle [radius=0.3] node (s2) {};
\draw[draw=none,fill=black!25!white] (1,0) circle [radius=0.24] node () {2};
\draw[thick, arrows={-Stealth[reversed,reversed]}] (0,0.3) to [bend left=45] (1,0.3);
\draw[thick, arrows={Stealth[reversed,reversed]-},dashed] (0,-0.3) to [bend right=45] (1,-0.3);
\draw[thick,fill=white] (2.2,0) circle [radius=0.3] node (sn) {N};
\node[] at (1.6,0) {...};
\draw[thick, arrows={-Stealth[reversed,reversed]}] (1,0.3) to [bend left=35] (2.2,0.3);
\draw[thick, arrows={Stealth[reversed,reversed]-},dashed] (1,-0.3) to [bend right=35] (2.2,-0.3);

\draw[thick, arrows={-Stealth[reversed,reversed]}] (2.2,0.3) to [bend left=35] (3.3,0.3);
\draw[thick, arrows={Stealth[reversed,reversed]-},dashed] (2.2,-0.3) to [bend right=35] (3.3,-0.3);

\draw[thick,fill=white] (3.3,0) circle [radius=0.3] node (sn1) {\footnotesize{N+1}};
\draw[thick, arrows={-Stealth[reversed,reversed]}] (3.3,0.3) to [bend left=45] (4.3,0.3);
\draw[thick, arrows={Stealth[reversed,reversed]-},dashed] (3.3,-0.3) to [bend right=45] (4.3,-0.3);
\draw[thick,fill=white] (4.3,0) circle [radius=0.3] node (sn2) {\footnotesize{N+2}};
\node[] at (4.9,0) {...};
\draw[thick, arrows={-Stealth[reversed,reversed]}] (4.3,0.3) to [bend left=35] (5.5,0.3);
\draw[thick, arrows={Stealth[reversed,reversed]-},dashed] (4.3,-0.3) to [bend right=35] (5.5,-0.3);
\draw[thick,fill=white] (5.5,0) circle [radius=0.3] node (s2n) {2N};
\draw[thick, arrows={-Stealth[reversed,reversed]}] (5.8,0.2) to [out=30,in=330,looseness=4] (5.8,-0.2);
\end{scope}

% notation
\begin{scope}[xshift=2.6cm,yshift=-1.2cm, scale=0.8, every node/.style={scale=0.8}]
\node[] at (1.95,-1.05) {\textcolor{dark_red}{\large{\textbf{Output=0}}}};
\node[] at (4.5,-1.05) {\textcolor{dark_green}{\large{\textbf{Output=1}}}};

\begin{scope}[xshift=-1.6cm,yshift=0cm]
\draw[rounded corners,densely dashed] (-1.2,-2.2) rectangle ++(8.7,0.8) node[] () {};
\draw[thick,fill=white] (-0.8,-1.8) circle [radius=0.22] node () {};
\node[] at (-0.12,-1.8) {State};
\draw[thick,fill=white] (0.8,-1.8) circle [radius=0.22] node () {};
\draw[draw=none,fill=black!25!white] (0.8,-1.8) circle [radius=0.17] node () {};
\node[] at (1.95,-1.8) {Active state};
\draw[thick, arrows={-Stealth[reversed,reversed]}] (3.05,-1.8) to (3.55,-1.8);
\node[] at (4.3,-1.8) {Increment};
\draw[thick, arrows={-Stealth[reversed,reversed]},dashed] (5.3,-1.8) to (5.8,-1.8);
\node[] at (6.65,-1.8) {Decrement};
\end{scope}

\end{scope}

\end{tikzpicture}
\vspace{-0.15cm}
\caption{Organization of TAs for a LUT.}
\label{fig:LUT}
\vspace{-0.4cm}
\end{figure}
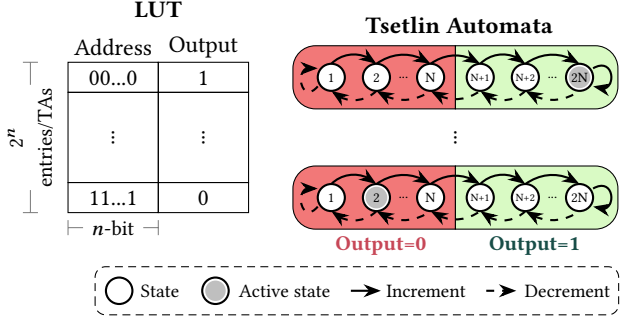

The training mechanism is as follows. 
The TA states for all LUTs are firstly randomly initialized to N or N+1, placing them near the point of maximum uncertainty, assuming a total of 2N states per TA.
The training is conducted per datapoint and over a fixed number of epochs. For each training sample, let $y$ denote the true label and $\hat{y}$ the label predicted by the model. If $y=\hat{y}$, the TA states across all LUTs remain unchanged. Feedback is triggered only when $y\neq\hat{y}$, and proceeds under a collective responsibility scheme:
\begin{itemize}[topsep=0pt, partopsep=0pt, itemsep=0pt, parsep=0pt]
	\item \textbf{Increment the correct discriminator $y$}: For all LUTs belonging to discriminator $y$, the TAs corresponding to the addresses determined by the training sample is incremented with a preset feedback probability $P$, where TA for each LUT makes an independent decision based on its own random draw. This potentially encourages more LUTs within the discriminator to support the correct decision, and enables diverse learning across LUTs.
	\item \textbf{Decrement the incorrect discriminator $\hat{y}$}: For all LUTs belonging to discriminator $\hat{y}$, the TAs corresponding to the addresses given by the training sample is decremented, potentially reducing the voting sum of the incorrect discriminator. Decrement for each LUT is similarly applied with $P$, independently across all LUTs.
	\item Discriminators other than $y$ and $\hat{y}$ undergo no change, as they do not affect the classification of the datapoint.
\end{itemize}

\subsection{Impact of Learning Parameters} \label{sec:param}
We investigate the impact of the following parameters on accuracy-number of TAs (determined by the number of LUTs and inputs per LUT), number of TA states, and feedback probability $P$-using experiments on the MNIST dataset \cite{deng2012mnist}.

\subsubsection{\textbf{Number of TAs}}
TA count increases linearly with LUTs. Figure~\ref{fig:accuracy_noLUTs} shows how training and test accuracy change with varying numbers of LUTs during training. Both accuracies improve as the number of LUTs increases, though the magnitude of improvement tend to decrease with each additional LUT.
\pgfplotsset{width=3cm}

% Style to select only points from #1 to #2 (inclusive)
\pgfplotsset{select coords between index/.style 2 args={
		x filter/.code={
			\ifnum\coordindex<#1\def\pgfmathresult{}\fi
			\ifnum\coordindex>#2\def\pgfmathresult{}\fi
		}
}}

    \newenvironment{customlegend}[1][]{%
	\begingroup
	\csname pgfplots@init@cleared@structures\endcsname
	\pgfplotsset{#1}%
}{%
	\csname pgfplots@createlegend\endcsname
	\endgroup
}%
\def\addlegendimage{\csname pgfplots@addlegendimage\endcsname}

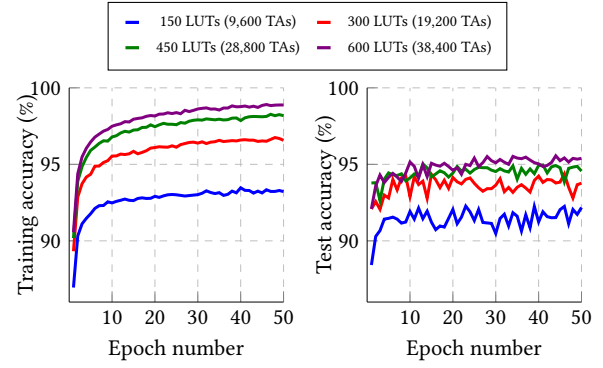
\begin{figure}[!htb]
\centering

\begin{minipage}{\linewidth}
	\centering
	\begin{tikzpicture}
	\draw[thin] (-4.3,0.1) rectangle (1,1);
	\begin{customlegend}[legend columns=2,legend style={align=left,draw=none,column sep=1ex, font=\scriptsize}, legend image post style={xscale=0.3},
	legend entries={{150 LUTs (9,600 TAs)},
		{300 LUTs (19,200 TAs)},
		{450 LUTs (28,800 TAs)},
		{600 LUTs (38,400 TAs)}		
	}]
	\addlegendimage{very thick, blue}
	\addlegendimage{very thick, red}
	\addlegendimage{very thick, green!50!black}
	\addlegendimage{very thick, violet}	   
	\end{customlegend}
	\end{tikzpicture}
\end{minipage}

		\begin{tikzpicture}[font=\normalsize, scale=0.95]
		\pgfplotsset{
			scale only axis,
		}
		
		\begin{axis}[
		height=3cm,
  	axis x line*=bottom,
	  axis y line*=left,
		ymax=100,
		ymin=86,
		xmin=0,
		xmax=50,
		xtick={10,20,30,40,50},
		grid=major,
		clip=false,
		grid style={dashed,gray!50},
		ylabel = Training accuracy (\%),
		xlabel = Epoch number,
        ylabel style={yshift=-2.5ex},
        legend style={at={(axis cs:-20, 105)},anchor=west,font=\scriptsize},
        legend image post style={xscale=0.3}]	
		\addplot [very thick, blue] table [y=Train_150, x expr=\thisrow{Epoch}+1] {Data/accuracy_noLUTs.dat};
		\addplot [very thick, red] table [y=Train_300, x expr=\thisrow{Epoch}+1] {Data/accuracy_noLUTs.dat};
		\addplot [very thick, green!50!black] table [y=Train_450, x expr=\thisrow{Epoch}+1] {Data/accuracy_noLUTs.dat};
		\addplot [very thick, violet] table [y=Train_600, x expr=\thisrow{Epoch}+1] {Data/accuracy_noLUTs.dat};			
		%\draw [thick,blue] (axis cs:1,91) -- node[]{} (axis cs:1,96);
		\end{axis}
	\end{tikzpicture}
		\begin{tikzpicture}[font=\normalsize, scale=0.95]
\pgfplotsset{
	scale only axis,
}

\begin{axis}[
height=3cm,
axis x line*=bottom,
axis y line*=left,
ymax=100,
ymin=86,
xmin=0,
xmax=50,
xtick={10,20,30,40,50},
grid=major,
grid style={dashed,gray!50},
ylabel = Test accuracy (\%),
xlabel = Epoch number,
ylabel style={yshift=-2.5ex},
clip=false]	
\addplot [very thick, blue] table [y=Test_150, x expr=\thisrow{Epoch}+1] {Data/accuracy_noLUTs.dat};
%\addlegendentry{150 LUTs (9,600 TAs)}
\addplot [very thick, red] table [y=Test_300, x expr=\thisrow{Epoch}+1] {Data/accuracy_noLUTs.dat};
%\addlegendentry{300 LUTs (19,200 TAs)}
\addplot [very thick, green!50!black] table [y=Test_450, x expr=\thisrow{Epoch}+1] {Data/accuracy_noLUTs.dat};
%\addlegendentry{450 LUTs (28,800 TAs)}
\addplot [very thick, violet] table [y=Test_600, x expr=\thisrow{Epoch}+1] {Data/accuracy_noLUTs.dat};
%\addlegendentry{600 LUTs (38,400 TAs)}			
\end{axis}
\end{tikzpicture}
\vspace{-0.4cm}
	\caption{Accuracy vs. LUTs per discriminator (6-input each).}
	\label{fig:accuracy_noLUTs}
\vspace{-0.4cm}
\end{figure}

TA count grows exponentially with inputs per LUT. Figure~\ref{fig:accuracy_noInputs} illustrates the impact of varying inputs per LUT, while keeping the total number of inputs to the model constant. Both training and test accuracy improve with more inputs per LUT. However, in practical FPGA implementations, the number of inputs per LUT is often fixed by the hardware, limiting flexibility in adjusting this parameter, especially for on-chip training.
\pgfplotsset{width=3cm}

% Style to select only points from #1 to #2 (inclusive)
\pgfplotsset{select coords between index/.style 2 args={
		x filter/.code={
			\ifnum\coordindex<#1\def\pgfmathresult{}\fi
			\ifnum\coordindex>#2\def\pgfmathresult{}\fi
		}
}}

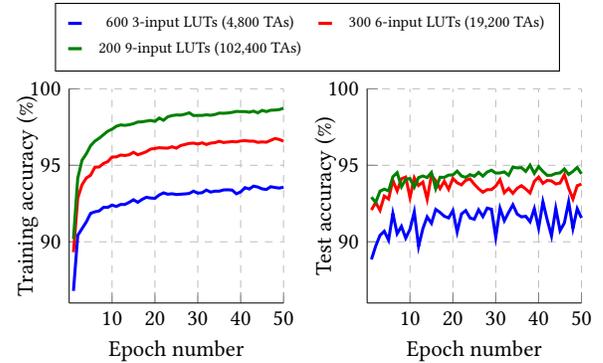
\begin{figure}[!htb]
\vspace{0cm}
\centering

\begin{minipage}{\linewidth}
	\centering
	\begin{tikzpicture}
	\draw[thin] (-5.7,0.1) rectangle (1,1);
	\begin{customlegend}[legend columns=2,legend style={align=left,draw=none,column sep=1ex, font=\scriptsize}, legend image post style={xscale=0.3},
	legend entries={
		{600 3-input LUTs (4,800 TAs)},
		{300 6-input LUTs (19,200 TAs)},
		{200 9-input LUTs (102,400 TAs)}		
	}]
	\addlegendimage{very thick, blue}
	\addlegendimage{very thick, red}
	\addlegendimage{very thick, green!50!black}   
	\end{customlegend}
	\end{tikzpicture}
\end{minipage}

		\begin{tikzpicture}[font=\normalsize, scale=0.95]
		\pgfplotsset{
			scale only axis,
		}
		
		\begin{axis}[
		height=3cm,
  	axis x line*=bottom,
	  axis y line*=left,
		ymax=100,
		ymin=86,
		xmin=0,
		xmax=50,
		xtick={10,20,30,40,50},
		grid=major,
		clip=false,
		grid style={dashed,gray!50},
		ylabel = Training accuracy (\%),
		xlabel = Epoch number,
        ylabel style={yshift=-2.5ex},
        legend style={at={(axis cs:-20, 105)},anchor=west,font=\scriptsize},
        legend image post style={xscale=0.3}]	
		\addplot [very thick, blue] table [y=Train_3in, x expr=\thisrow{Epoch}+1] {Data/accuracy_noInputs.dat};
		\addplot [very thick, red] table [y=Train_6in, x expr=\thisrow{Epoch}+1] {Data/accuracy_noInputs.dat};
		\addplot [very thick, green!50!black] table [y=Train_9in, x expr=\thisrow{Epoch}+1] {Data/accuracy_noInputs.dat};			
		%\draw [thick,blue] (axis cs:1,91) -- node[]{} (axis cs:1,96);
		\end{axis}
	\end{tikzpicture}
		\begin{tikzpicture}[font=\normalsize, scale=0.95]
\pgfplotsset{
	scale only axis,
}

\begin{axis}[
height=3cm,
axis x line*=bottom,
axis y line*=left,
ymax=100,
ymin=86,
xmin=0,
xmax=50,
xtick={10,20,30,40,50},
grid=major,
grid style={dashed,gray!50},
ylabel = Test accuracy (\%),
xlabel = Epoch number,
ylabel style={yshift=-2.5ex},
clip=false]	
\addplot [very thick, blue] table [y=Test_3in, x expr=\thisrow{Epoch}+1] {Data/accuracy_noInputs.dat};
\addplot [very thick, red] table [y=Test_6in, x expr=\thisrow{Epoch}+1] {Data/accuracy_noInputs.dat};
\addplot [very thick, green!50!black] table [y=Test_9in, x expr=\thisrow{Epoch}+1] {Data/accuracy_noInputs.dat};		
\end{axis}
\end{tikzpicture}
\vspace{-0.4cm}
	\caption{Accuracy vs. inputs per LUT.}
	\label{fig:accuracy_noInputs}
\vspace{-0.4cm}
\end{figure}

Overall, accuracy improves with increased resources, as more TAs allow additional LUT entries to capture potential patterns. In practice, this enables adaptive resource allocation—particularly by adjusting the number of LUTs—and model scaling to balance accuracy against resource usage.

\subsubsection{\textbf{Number of TA states}}
\begin{figure}[!htb]
%\vspace{-0.2cm}
	\centering
	
\begin{tikzpicture}
\begin{axis}[
width=4.4cm,
height=4cm,
ybar,
enlargelimits=0.2,
ylabel={\begin{tabular}{@{}c@{}} Highest training \\ accuracy (\%) \end{tabular}},
xlabel={TA states},
symbolic x coords={32,64,128,256},
xtick=data,
nodes near coords,
nodes near coords align={vertical},
every node near coord/.style={font=\scriptsize},
ylabel style={yshift=-2ex},
font=\small
]
\addplot[blue, fill=babyblue] coordinates {(32,94.47) (64,96.06) (128,96.75) (256,96.75)};
\end{axis}
\end{tikzpicture}
\begin{tikzpicture}
\begin{axis}[
width=4.4cm,
height=4cm,
ybar,
enlargelimits=0.2,
ylabel={\begin{tabular}{@{}c@{}} Highest test \\ accuracy (\%) \end{tabular}},
xlabel={TA states},
symbolic x coords={32,64,128,256},
xtick=data,
nodes near coords,
nodes near coords align={vertical},
every node near coord/.style={font=\scriptsize},
ylabel style={yshift=-2ex},
ymin=93,
ymax=95,
font=\small
]
\addplot[red, fill=mypink!50!white] coordinates {(32,93.62) (64,94.22) (128,94.41) (256,94.38)};
\end{axis}
\end{tikzpicture}
\vspace{-0.3cm}
	\caption{Best accuracy over 50 epochs with varying TA states (300 6-input LUTs per class).}
	\label{fig:accuracy_TAstates}
\vspace{-0.3cm}
\end{figure}
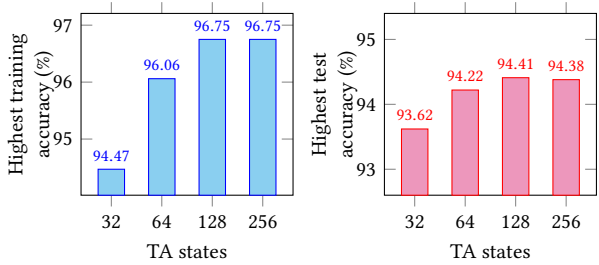

In Figure~\ref{fig:accuracy_TAstates}, both training and test accuracy increase as the number of TA states grows from a small value, but generally stabilize once the number of states exceeds 128. The number of TA states controls the resolution of a TA’s decision to flip between 0 and 1 for a LUT entry. While more states improve robustness to noise by making TAs less sensitive to occasional feedback, this effect saturates once enough states are available to resist noise-induced fluctuations.

\subsubsection{\textbf{Feedback probability}}
The feedback is probabilistically triggered with probability $P$ to avoid the training process being trapped in local optima. As shown in Figure~\ref{fig:accuracy_TrainProb}, TsetlinWiSARD consistently converges across different values of $P$, with higher $P$ generally leading to faster convergence.

\pgfplotsset{width=5cm}

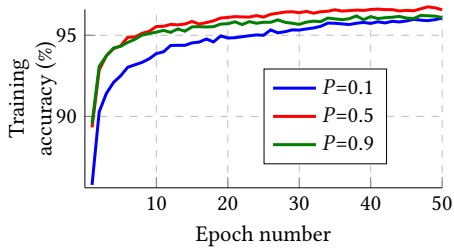
\begin{figure}[!htb]
\centering
%\vspace{-0.3cm}
		\begin{tikzpicture}[font=\normalsize, scale=0.95]
		\pgfplotsset{
			scale only axis,
		}
		
		\begin{axis}[
		height=2.5cm,
  	axis x line*=bottom,
	  axis y line*=left,
		ymax=96.6,
		ymin=85.6,
		xmin=0,
		xmax=50,
		xtick={10,20,30,40,50},
		grid=major,
		clip=false,
		grid style={dashed,gray!50},
		ylabel = \begin{tabular}{@{}c@{}} Training \\ accuracy (\%) \end{tabular},
		xlabel = Epoch number,
        ylabel style={yshift=-2ex},
        legend style={at={(axis cs:25, 90)},anchor=west}]	
		\addplot [very thick, blue] table [y=Train_01, x expr=\thisrow{Epoch}+1] {Data/accuracy_prob.dat};
		\addlegendentry{$P$=0.1}
		\addplot [very thick, red] table [y=Train_05, x expr=\thisrow{Epoch}+1] {Data/accuracy_prob.dat};
		\addlegendentry{$P$=0.5}
		\addplot [very thick, green!50!black] table [y=Train_09, x expr=\thisrow{Epoch}+1] {Data/accuracy_prob.dat};
		\addlegendentry{$P$=0.9}
		%\draw [thick,blue] (axis cs:1,91) -- node[]{} (axis cs:1,96);
		\end{axis}
	\end{tikzpicture}
\vspace{-0.4cm}	
	\caption{Training accuracy vs. feedback probability $P$.}
	\label{fig:accuracy_TrainProb}
\vspace{-0.5cm}
\end{figure}

This convergence can be explained as follows: 

Consider a feature subset that supports the correct classification of a given class and indexes a specific LUT entry. Over time, even with a small $P$, the probability of receiving at least one reinforcement becomes effectively 1 as the number of training samples grows large (1-(1-$P)^n\rightarrow1$, if $n\rightarrow\infty$). Therefore, with enough feedback opportunities, the corresponding TA state will eventually be incremented, setting the LUT entry to 1 and enabling it to contribute a vote during inference. Similarly, feature subsets that oppose a given class will receive decremental feedback and their entries will eventually be pushed to 0.

Consider a feature subset irrelevant to any class ($i.e.$, noise) that indexes a LUT entry. The corresponding TA will be randomly incremented or decremented with equal probability $P$, effectively behaving as a finite, irreducible, and aperiodic Markov chain. Over time, its active state distribution becomes uniform and independent of $P$ \cite{plavnick2008fundamental}. Since this feature subset is shared across all discriminators, its contributions are evenly distributed, introducing no bias and thus preserving classification robustness against irrelevant features.

We further evaluate test accuracy under varying values of $P$, as shown in Figure~\ref{fig:accuracy_Prob}. Notably, a small $P$ ($e.g.$, $P$=0.1) leads to a steady increase in test accuracy, suggesting more stable progress toward a global optimum, while larger values of $P$ tend to introduce greater fluctuations during training. Nevertheless, all tested values of $P$ ultimately reach similar accuracy levels, given sufficient training epochs. This indicates that TsetlinWiSARD is generally insensitive to the choice of $P$, a key observation that guides the design of our compact pseudo-random number generator discussed in Section~\ref{sec:design}.

\pgfplotsset{width=2.3cm}

% Style to select only points from #1 to #2 (inclusive)
\pgfplotsset{select coords between index/.style 2 args={
		x filter/.code={
			\ifnum\coordindex<#1\def\pgfmathresult{}\fi
			\ifnum\coordindex>#2\def\pgfmathresult{}\fi
		}
}}

\begin{figure}[!htb]
    %\vspace{-0.2cm}

\centering
\begin{tikzpicture}[font=\normalsize, scale=0.95]
\pgfplotsset{
	scale only axis,
}

\begin{axis}[
height=2.5cm,
axis x line*=bottom,
axis y line*=left,
ymax=95,
ymin=88,
xmin=0,
xmax=50,
xtick={10,20,30,40,50},
grid=major,
grid style={dashed,gray!50},
ylabel = Test accuracy (\%),
xlabel = Epoch number,
ylabel style={yshift=-1ex},
clip=false
]	
\addplot [very thick, blue] table [y=Test_01, x expr=\thisrow{Epoch}+1] {Data/accuracy_prob.dat};
\draw [gray, dashed] (axis cs: 0,90) -- (axis cs: 58,90);
\draw [gray, dashed] (axis cs: 0,92) -- (axis cs: 58,92);
\draw [gray, dashed] (axis cs: 0,94) -- (axis cs: 58,94);
\node[] at (axis cs: 32,91) {\textbf{$P$=0.1}};		
\end{axis}
\end{tikzpicture}
\begin{tikzpicture}[font=\normalsize, scale=0.95]
\pgfplotsset{
	scale only axis,
}

\begin{axis}[
height=2.5cm,
axis x line*=bottom,
axis y line*=left,
ymax=95,
ymin=88,
xmin=0,
xmax=50,
xtick={10,20,30,40,50},
grid=major,
grid style={dashed,gray!50},
xlabel = Epoch number,
ytick=\empty,      % <-- Removes y ticks
yticklabels=\empty,
clip=false
]	
\addplot [very thick, red] table [y=Test_05, x expr=\thisrow{Epoch}+1] {Data/accuracy_prob.dat};
\draw [gray, dashed] (axis cs: 0,90) -- (axis cs: 56,90);
\draw [gray, dashed] (axis cs: 0,92) -- (axis cs: 56,92);
\draw [gray, dashed] (axis cs: 0,94) -- (axis cs: 56,94);
\node[] at (axis cs: 32,91) {\textbf{$P$=0.5}};		
\end{axis}
\end{tikzpicture}
\begin{tikzpicture}[font=\normalsize, scale=0.95]
\pgfplotsset{
	scale only axis,
}

\begin{axis}[
height=2.5cm,
axis x line*=bottom,
axis y line*=left,
ymax=95,
ymin=88,
xmin=0,
xmax=50,
xtick={10,20,30,40,50},
grid=major,
grid style={dashed,gray!50},
xlabel = Epoch number,
ytick=\empty,      % <-- Removes y ticks
yticklabels=\empty
]	
\addplot [very thick, green!50!black] table [y=Test_09, x expr=\thisrow{Epoch}+1] {Data/accuracy_prob.dat};
\draw [gray, dashed] (axis cs: 0,90) -- (axis cs: 58,90);
\draw [gray, dashed] (axis cs: 0,92) -- (axis cs: 58,92);
\draw [gray, dashed] (axis cs: 0,94) -- (axis cs: 58,94);
\node[] at (axis cs: 32,91) {\textbf{$P$=0.9}};			
\end{axis}
\end{tikzpicture}
	\caption{Test accuracy vs. feedback probability $P$.}
	\label{fig:accuracy_Prob}
    \vspace{-0.6cm}
\end{figure}
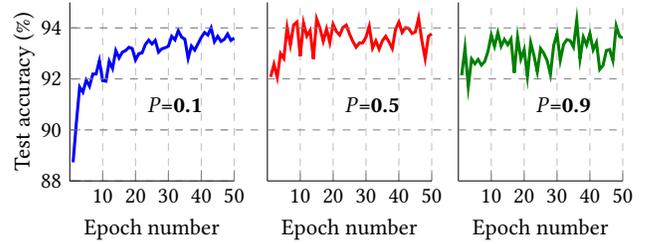

\subsection{On-Chip Training Architecture on FPGA} \label{sec:design}
Modern FPGAs support LUTs in a writeable configuration, enabling them to function as distributed RAMs—such as LUTRAMs in Xilinx and MLABs in Intel (Altera) devices. This allows LUTs to act as small, dynamically writable memory blocks at runtime. In our on-chip training architecture, both the logical LUTs and TAs in TsetlinWiSARD are mapped to these physical distributed RAMs.

Leveraging the fact that only one entry per LUT is potentially updated during each feedback cycle, TA states for all entries in a LUT—referred to as a TA team—can be compactly stored in a small number of distributed RAMs. As shown in Figure~\ref{fig:architecture}(a), we use Xilinx 64$\times$1 LUTRAMs to store the TA states in binary. To represent 2N TA states, we use $log_2($N)+1 LUTRAMs per LUT. All LUTRAMs in a TA team share the same feature subset as the address input and are each paired with a full adder (FA), forming a ripple-carry adder to increment or decrement the TA states during training. The \textit{feedback\_en} signal controls write operations—enabling state updates during training when high, while disabling writes during inference. For classification, only the most significant bit (MSB) of each TA state is used to determine the binary output of each entry—indicating whether the state lies above or below the midpoint. 

\input{src/architecture}

Figure~\ref{fig:architecture}(b) illustrates the overall TsetlinWiSARD architecture. All LUTs within a discriminator are controlled by $feedback$ and \textit{feedback\_dir}. Please refer to Section~\ref{sec:training} for details of the training mechanism. A linear feedback shift register (LFSR) serves as a compact pseudo-random number generator to probabilistically enable feedback. An $m$-stage LFSR is used for $m$ LUTs, with each stage independently controlling feedback to its corresponding LUT, typically with a 0.5 probability, as the internal stages of a LFSR naturally exhibit a 0.5 probability. As shown in Section~\ref{sec:param}, the choice of feedback probability has little impact on convergence or accuracy, making the bitwise LFSR a lightweight and effective solution. Each discriminator requires at least one LFSR, determined by the number of LUTs per discriminator, and each LFSR is assigned a unique initial value to promote diversity. 
\vspace{-0.2cm}
\section{Off-Chip Training Results} \label{sec:off-chip}
We first evaluate the performance of TsetlinWiSARD in terms of classification accuracy through offline training on six benchmark datasets: MNIST \cite{deng2012mnist}, Fashion-MNIST (FMNIST) \cite{xiao2017fashion}, Kuzushiji-MNIST (KMNIST) \cite{clanuwat2018deep}, electromyography (EMG)-based gesture recognition \cite{lobov2018latent}, human activity recognition (HAR) \cite{anguita2013public}, and gesture phase segmentation (GPS) \cite{madeo2013gesture}. These datasets have been preprocessed via thresholding to produce Boolean feature vectors of size 784, 784, 784, 160, 560, and 180, respectively.
The preprocessed datasets, along with the source code for TsetlinWiSARD off-chip training, are publicly available at: \url{https://github.com/nsd5g13/TsetlinWiSARD}.

\begin{figure}[!htb]
%\vspace{-0.3cm}
	\centering
\subfloat[]{	
\begin{tikzpicture}[font=\normalsize]

\begin{axis}[
%x tick label style={/pgf/number format/1000 sep=},
width=8.5cm,
height=4cm,
ybar,
ymin=0, 
ymax=100,
ytick={0,50,100},
bar width=0.25cm, %width=1\textwidth,
legend style={at={(0.9,1.3)}, anchor=north east, legend columns=-1, font=\footnotesize},
ylabel={Test accuracy (\%)},
xticklabels from table={\datatableS}{dataset},
xtick=data,
ylabel style={yshift=-2ex},
ymajorgrids,
clip = false,
%grid style={dashed,gray!50},
%x tick label style={rotate=45,anchor=east},
nodes near coords,
node near coords style={yshift=-1cm,
	/utils/exec={\setbox0\hbox{\pgfmathprintnumber\pgfplotspointmeta}
		\pgfmathfloattomacro{\pgfplotspointmeta}{\F}{\M}{\E}
		\pgfmathsetmacro{\myanchor}{ifthenelse(\M*pow(10,\E-3)*2-\the\wd0>0,"east","west")}
		%\typeout{\M,\E,\the\wd0,\mysign,\myanchor}
	},
	rotate=90,anchor=\myanchor,font=\scriptsize,
	/pgf/number format/.cd,fixed zerofill,precision=2},
]

\pgfplotsinvokeforeach {1,...,3}{
	\addplot table [x expr=\coordindex, y index=#1] {\datatableS};
\draw[densely dotted, thick] (axis cs: 0.5,0) -- (axis cs: 0.5,100);
\draw[densely dotted, thick] (axis cs: 1.5,0) -- (axis cs: 1.5,100);
\draw[densely dotted, thick] (axis cs: 2.5,0) -- (axis cs: 2.5,100);
\draw[densely dotted, thick] (axis cs: 3.5,0) -- (axis cs: 3.5,100);
\draw[densely dotted, thick] (axis cs: 4.5,0) -- (axis cs: 4.5,100);
}
\legend{WiSARD, B-bleaching, TsetlinWiSARD}
\end{axis}
\end{tikzpicture}
}
\vspace{-0.4cm}
\subfloat[]{
\begin{tikzpicture}[font=\normalsize]

\begin{axis}[
%x tick label style={/pgf/number format/1000 sep=},
width=8.5cm,
height=4cm,
ybar,
ymin=0, 
ymax=100,
ytick={0,50,100},
bar width=0.25cm, %width=1\textwidth,
legend style={at={(1,1)}, anchor=north east, legend columns=-1, font=\footnotesize},
ylabel={Test accuracy (\%)},
xticklabels from table={\datatableL}{dataset},
xtick=data,
ylabel style={yshift=-2ex},
ymajorgrids,
%grid=major,
%grid style={dashed,gray!50},
%x tick label style={rotate=45,anchor=east},
nodes near coords,
node near coords style={yshift=-1cm,
	/utils/exec={\setbox0\hbox{\pgfmathprintnumber\pgfplotspointmeta}
		\pgfmathfloattomacro{\pgfplotspointmeta}{\F}{\M}{\E}
		\pgfmathsetmacro{\myanchor}{ifthenelse(\M*pow(10,\E-3)*2-\the\wd0>0,"east","west")}
		%\typeout{\M,\E,\the\wd0,\mysign,\myanchor}
	},
	rotate=90,anchor=\myanchor,font=\scriptsize,
	/pgf/number format/.cd,fixed zerofill,precision=2},
]

\pgfplotsinvokeforeach {1,...,3}{
	\addplot table [x expr=\coordindex, y index=#1] {\datatableL};
\draw[densely dotted, thick] (axis cs: 0.5,0) -- (axis cs: 0.5,100);
\draw[densely dotted, thick] (axis cs: 1.5,0) -- (axis cs: 1.5,100);
\draw[densely dotted, thick] (axis cs: 2.5,0) -- (axis cs: 2.5,100);
\draw[densely dotted, thick] (axis cs: 3.5,0) -- (axis cs: 3.5,100);
\draw[densely dotted, thick] (axis cs: 4.5,0) -- (axis cs: 4.5,100);
}
%\legend{WiSARD, B-WiSARD, TsetlinWiSARD}
\end{axis}
\end{tikzpicture}
}
\vspace{-0.4cm}
	\caption{Accuracy comparison of different training methods for WiSARD with (a) 150 and (b) 300 LUTs per discriminator.}
\label{fig:results_WNNs}
\vspace{-0.3cm}
\end{figure}
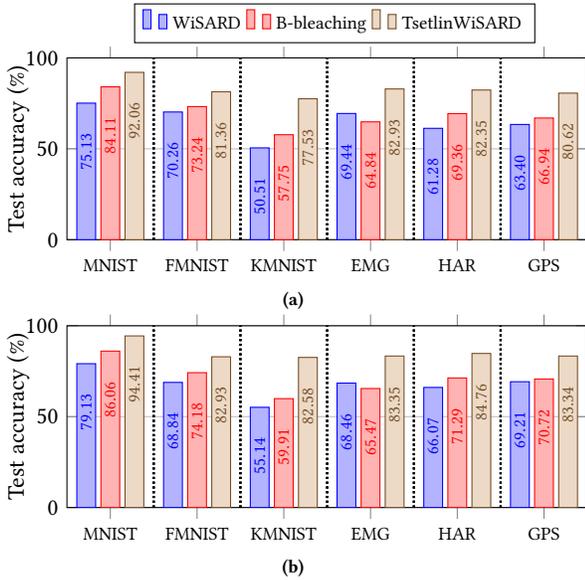

We compare accuracy of TsetlinWiSARD with that of standard WiSARD \cite{aleksander1984wisard} and B-bleaching \cite{bleaching}, using identical WiSARD architectures with 150 and 300 LUTs per discriminator.
For standard WiSARD, training continues until validation accuracy begins to decline, to prevent overfitting. For B-bleaching, the threshold is selected to achieve the highest accuracy. TsetlinWiSARD models are trained for a fixed 50 epochs.
As shown in Figure~\ref{fig:results_WNNs}, TsetlinWiSARD outperforms both training approaches across all configurations.

\pgfplotstableread[col sep=comma]{
	
	dataset,       WiSARD,     B-WiSARD,    TsetlinWiSARD
	{\footnotesize{MNIST}},     75.13,     84.11,      92.06
	{\footnotesize{FMNIST}},    70.26,     73.24,      81.36
	{\footnotesize{KMNIST}},    50.51,     57.75,      77.53
	{\footnotesize{EMG}},    69.44,      64.84,      82.93
	{\footnotesize{HAR}},    61.28,      69.36,      82.35
	{\footnotesize{GPS}},    63.40,      66.94,      80.62
	
}\datatableS

\pgfplotstableread[col sep=comma]{
	
	dataset,       WiSARD,     B-WiSARD,    TsetlinWiSARD
	{\footnotesize{MNIST}},     79.13,     86.06,      94.41
	{\footnotesize{FMNIST}},    68.84,     74.18,      82.93
	{\footnotesize{KMNIST}},    55.14,     59.91,      82.58
	{\footnotesize{EMG}},    68.46,      65.47,      83.35
	{\footnotesize{HAR}},    66.07,      71.29,      84.76
	{\footnotesize{GPS}},    69.21,      70.72,      83.34
	
}\datatableL

We further compare TsetlinWiSARD with the other TA-based ML algorithm, TM, where both models benefit from increased numbers of TAs. To ensure a fair comparison, we vary the number of TAs in both models—by adjusting the number of clauses in TM and the number of 6-input LUTs in TsetlinWiSARD—and report their accuracy in Figure~\ref{fig:WNN_vs_TM}. 
Both models are trained for 50 epochs on each dataset.
Overall, TsetlinWiSARD achieves comparable or superior accuracy using more than one-third fewer TAs across all datasets except EMG, where it still shows more than one-third TA reduction. This efficiency stems from the architectural difference: TM allocates one TA per input (and its complement) per clause, requiring significantly more TAs for high-dimensional data; in contrast, TsetlinWiSARD assigns one TA to each LUT entry, enabling pattern learning over subsets of features using fewer TAs. Additionally, TA count alone does not directly translate to hardware cost. As shown in Figure~\ref{fig:architecture}, TsetlinWiSARD can compactly store large numbers of TAs using a few LUTRAMs, since only one TA per LUT may be updated per feedback. In contrast, TM may update multiple TAs per feedback cycle, leading to higher on-chip resource usage, since TA resources cannot be shared across simultaneous updates.

The sustained improvements in accuracy compared to earlier WiSARD training and TMs, observed across multiple datasets and model sizes, highlight the scalability of TsetlinWiSARD.

\pgfplotsset{width=2cm}

% index 2: 0-TWisard, 1-TM
\pgfplotstableread[col sep=comma,header=false]{
0.96, 92.18, 0
1.92, 94.38, 0
2.88, 95.03, 0
0.9405, 85.15, 1
1.8816, 88.47, 1 
2.8224, 91.65, 1
}\mnist

\pgfplotstableread[col sep=comma,header=false]{
0.9600, 81.36, 0
1.9200, 82.93, 0
2.8800, 83.90, 0
0.9405, 75.31, 1
1.8816, 77.39, 1 
2.8224, 79.06, 1
}\fmnist

\pgfplotstableread[col sep=comma,header=false]{
	0.9600, 77.53, 0
	1.9200, 82.58, 0
	2.8800, 84.69, 0
	0.9405, 52.48, 1
	1.8816, 61.01, 1 
	2.8224, 65.16, 1
}\kmnist

\pgfplotstableread[col sep=comma,header=false]{
	3.200, 81.66, 0
	6.400, 83.14, 0
	9.600, 82.93, 0
	3.200, 80.61, 1
	6.400, 81.81, 1
	9.600, 83.21, 1
}\emg

\pgfplotstableread[col sep=comma,header=false]{
	0.9600, 82.35, 0
	1.9200, 84.76, 0
	2.8800, 85.75, 0
	0.8960, 79.06, 1
	2.0160, 83.54, 1 
	2.9120, 83.17, 1
}\har

\pgfplotstableread[col sep=comma,header=false]{
	3.200, 70.92, 0
	6.400, 76.48, 0
	9.600, 80.62, 0
	2.88, 59.16, 1
	6.48, 63.3, 1 
	9.36, 64.97, 1
}\gps

\begin{figure}[!htb]
\vspace{-0.6cm}
	\centering
\subfloat{	
\begin{minipage}{\linewidth}
	\centering
	\begin{tikzpicture}[baseline=(current bounding box.north)]
	\draw[thin] (-1.95,0.48) rectangle (1,1);
	\begin{customlegend}[legend columns=2,legend style={align=left,draw=none,column sep=1ex, font=\scriptsize}, legend image post style={scale=1},
	legend entries={
		{TsetlinWiSARD},
		{TM}	
	}]
	\addlegendimage{only marks, mark=ball, ball color=babyblue, blue}
	\addlegendimage{only marks, mark=ball,ball color=mypink, red}
	\end{customlegend}
	\end{tikzpicture}
\end{minipage}
}
\vspace{-0.4cm}
\setcounter{subfigure}{0}
\subfloat[]{% %mnist
	\begin{tikzpicture}[font=\normalsize, scale=0.95]
	\pgfplotsset{
		scale only axis,
	}
	
	\begin{axis}[
	height=2.5cm,
	axis x line*=bottom,
	axis y line*=left,
	grid=major,
	grid style={dashed,gray!50},
	ylabel = Test accuracy (\%),
	xlabel = Number of TAs ($\times 10^4$),
	xlabel style={font=\scriptsize,yshift=1ex},
	ylabel style={yshift=-1ex},
	clip=false
	%x tick label style={
	%	/pgf/number format/.cd,
	%		fixed,
	%	fixed zerofill,
	%	precision=1,
	%	sci,
	%	/tikz/.cd,
	%	font=\tiny
	%},scaled x ticks=false
	]	
\addplot+[scatter, only marks,
scatter/classes={0={mark=ball, ball color=babyblue, blue, scale=1.5},
	1={mark=ball,ball color=mypink, red, scale=1.5}
},
scatter src=explicit symbolic
] table[x index=0,y index=1,meta index=2] \mnist;
\node[] at (axis cs: 1.0400,93.5) {\textcolor{blue}{\scriptsize 150L}};
\node[] at (axis cs: 1.9200, 93) {\textcolor{blue}{\scriptsize 300L}};
\node[] at (axis cs: 2.8200, 93.7) {\textcolor{blue}{\scriptsize 450L}};
\node[] at (axis cs: 1.0400,86.4) {\textcolor{red}{\scriptsize 6C}};
\node[] at (axis cs: 1.9200, 87.1) {\textcolor{red}{\scriptsize 12C}};
\node[] at (axis cs: 2.8200, 90.3) {\textcolor{red}{\scriptsize 18C}};
\end{axis}
	\end{tikzpicture}%
}\hspace{-0.1cm}
\subfloat[]{% %fmnist
	\begin{tikzpicture}[font=\normalsize, scale=0.95]
	\pgfplotsset{
		scale only axis,
	}
	
	\begin{axis}[
	height=2.5cm,
	axis x line*=bottom,
	axis y line*=left,
	grid=major,
	grid style={dashed,gray!50},
	%ylabel = Test accuracy (\%),
	xlabel = Number of TAs ($\times 10^4$),
xlabel style={font=\scriptsize,yshift=1ex},
	clip=false
	]	
	\addplot+[scatter, only marks,
	scatter/classes={0={mark=ball, ball color=babyblue, blue, scale=1.5},
		1={mark=ball,ball color=mypink, red, scale=1.5}
	},
	scatter src=explicit symbolic
	] table[x index=0,y index=1,meta index=2] \fmnist;
	\node[] at (axis cs: 1.0600,82.6) {\textcolor{blue}{\scriptsize 150L}};
	\node[] at (axis cs: 1.9200, 81.9) {\textcolor{blue}{\scriptsize 300L}};
	\node[] at (axis cs: 2.8200, 82.9) {\textcolor{blue}{\scriptsize 450L}};
	\node[] at (axis cs: 1.0100,76.4) {\textcolor{red}{\scriptsize 6C}};
	\node[] at (axis cs: 1.9200, 76.4) {\textcolor{red}{\scriptsize 12C}};
	\node[] at (axis cs: 2.8200, 78.1) {\textcolor{red}{\scriptsize 18C}};
	\end{axis}
	\end{tikzpicture}%
}\hspace{-0.1cm}
\subfloat[]{% %kmnist
	\begin{tikzpicture}[font=\normalsize, scale=0.95]
	\pgfplotsset{
		scale only axis,
	}
	
	\begin{axis}[
	height=2.5cm,
	axis x line*=bottom,
	axis y line*=left,
	grid=major,
	grid style={dashed,gray!50},
	%ylabel = Test accuracy (\%),
	xlabel = Number of TAs ($\times 10^4$),
xlabel style={font=\scriptsize,yshift=1ex},
clip=false
	]	
	\addplot+[scatter, only marks,
	scatter/classes={0={mark=ball, ball color=babyblue, blue, scale=1.5},
		1={mark=ball,ball color=mypink, red, scale=1.5}
	},
	scatter src=explicit symbolic
	] table[x index=0,y index=1,meta index=2] \kmnist;
	\node[] at (axis cs: 1.0600,82.5) {\textcolor{blue}{\scriptsize 150L}};
	\node[] at (axis cs: 1.9200, 87) {\textcolor{blue}{\scriptsize 300L}};
	\node[] at (axis cs: 2.8200, 80) {\textcolor{blue}{\scriptsize 450L}};
	\node[] at (axis cs: 1.0100,57) {\textcolor{red}{\scriptsize 6C}};
	\node[] at (axis cs: 1.9200, 56) {\textcolor{red}{\scriptsize 12C}};
	\node[] at (axis cs: 2.8200, 60) {\textcolor{red}{\scriptsize 18C}};
	\end{axis}
	\end{tikzpicture}%
}
\vspace{-0.5cm}
\subfloat[]{% %emg
	\begin{tikzpicture}[font=\normalsize, scale=0.95]
	\pgfplotsset{
		scale only axis,
	}
	
	\begin{axis}[
	height=2.5cm,
	axis x line*=bottom,
	axis y line*=left,
	xtick={3, 6, 9},
	grid=major,
	grid style={dashed,gray!50},
	ylabel = Test accuracy (\%),
	xlabel = Number of TAs ($\times 10^3$),
xlabel style={font=\scriptsize,yshift=1ex},
ylabel style={yshift=-1ex},
clip=false
	]	
	\addplot+[scatter, only marks,
	scatter/classes={0={mark=ball, ball color=babyblue, blue, scale=1.5},
		1={mark=ball,ball color=mypink, red, scale=1.5}
	},
	scatter src=explicit symbolic
	] table[x index=0,y index=1,meta index=2] \emg;
	\node[] at (axis cs: 3.300,82) {\textcolor{blue}{\scriptsize 50L}};
	\node[] at (axis cs: 6.400, 82.8) {\textcolor{blue}{\scriptsize 100L}};
	\node[] at (axis cs: 9.000, 82.6) {\textcolor{blue}{\scriptsize 150L}};
	\node[] at (axis cs: 4.400,80.5) {\textcolor{red}{\scriptsize 10C}};
	\node[] at (axis cs: 6.100, 81.5) {\textcolor{red}{\scriptsize 20C}};
	\node[] at (axis cs: 9.000, 83.5) {\textcolor{red}{\scriptsize 30C}};
	\end{axis}
	\end{tikzpicture}%
}\hspace{-0.1cm}
\subfloat[]{% %har
	\begin{tikzpicture}[font=\normalsize, scale=0.95]
	\pgfplotsset{
		scale only axis,
	}
	
	\begin{axis}[
	height=2.5cm,
	axis x line*=bottom,
	axis y line*=left,
	grid=major,
	grid style={dashed,gray!50},
    xtick={1, 2, 3},
	%ylabel = Test accuracy (\%),
	xlabel = Number of TAs ($\times 10^4$),
    xlabel style={font=\scriptsize,yshift=1ex},
	ylabel style={yshift=-1ex},
	clip=false
	%x tick label style={
	%	/pgf/number format/.cd,
	%	fixed,
	%	fixed zerofill,
	%	precision=1,
	%	sci,
	%	/tikz/.cd,
	%	font=\scriptsize
	%},scaled x ticks=false,
	]	
	\addplot+[scatter, only marks,
	scatter/classes={0={mark=ball, ball color=babyblue, blue, scale=1.5},
		1={mark=ball,ball color=mypink, red, scale=1.5}
	},
	scatter src=explicit symbolic
	] table[x index=0,y index=1,meta index=2] \har;
	\node[] at (axis cs: 1.0100,83.5) {\textcolor{blue}{\scriptsize 150L}};
	\node[] at (axis cs: 1.9200, 85.5) {\textcolor{blue}{\scriptsize 300L}};
	\node[] at (axis cs: 2.8200, 86.5) {\textcolor{blue}{\scriptsize 450L}};
	\node[] at (axis cs: 1.2500,79) {\textcolor{red}{\scriptsize 8C}};
	\node[] at (axis cs: 1.9200, 82.5) {\textcolor{red}{\scriptsize 18C}};
	\node[] at (axis cs: 2.8200, 82.5) {\textcolor{red}{\scriptsize 26C}};
	\end{axis}
	\end{tikzpicture}%
}\hspace{-0.1cm}
\subfloat[]{% %gps
	\begin{tikzpicture}[font=\normalsize, scale=0.95]
	\pgfplotsset{
		scale only axis,
	}
	
	\begin{axis}[
	height=2.5cm,
	axis x line*=bottom,
	axis y line*=left,
	grid=major,
	grid style={dashed,gray!50},
	%ylabel = Test accuracy (\%),
	xtick={3, 6, 9},
	grid=major,
	grid style={dashed,gray!50},
	%ylabel = Test accuracy (\%),
	xlabel = Number of TAs ($\times 10^3$),
xlabel style={font=\scriptsize,yshift=1ex},
ylabel style={yshift=-3ex},
clip=false
	%x tick label style={
	%	/pgf/number format/.cd,
	%	fixed,
	%	fixed zerofill,
	%	precision=1,
	%	sci,
	%	/tikz/.cd,
	%	font=\scriptsize
	%},scaled x ticks=false,
	]	
	\addplot+[scatter, only marks,
	scatter/classes={0={mark=ball, ball color=babyblue, blue, scale=1.5},
		1={mark=ball,ball color=mypink, red, scale=1.5}
	},
	scatter src=explicit symbolic
	] table[x index=0,y index=1,meta index=2] \gps;
	\node[] at (axis cs: 3.300,73) {\textcolor{blue}{\scriptsize 50L}};
	\node[] at (axis cs: 6.400, 79) {\textcolor{blue}{\scriptsize 100L}};
	\node[] at (axis cs: 9.000, 78) {\textcolor{blue}{\scriptsize 150L}};
	\node[] at (axis cs: 4.00,60) {\textcolor{red}{\scriptsize 8C}};
	\node[] at (axis cs: 6.100, 66) {\textcolor{red}{\scriptsize 18C}};
	\node[] at (axis cs: 9.000, 68) {\textcolor{red}{\scriptsize 26C}};
	\end{axis}
	\end{tikzpicture}%
}
\vspace{-0.4cm}
	\caption{Accuracy vs. TA counts for TsetlinWiSARD and TMs on (a) MNIST, (b) FMNIST, (c) KMNIST, (d) EMG, (e) HAR, and (f) GPS. The numbers before ``L" and ``C" represent LUT and clause counts in TsetlinWiSARD and TMs, respectively.}
\label{fig:WNN_vs_TM}
\vspace{-0.3cm}
\end{figure}
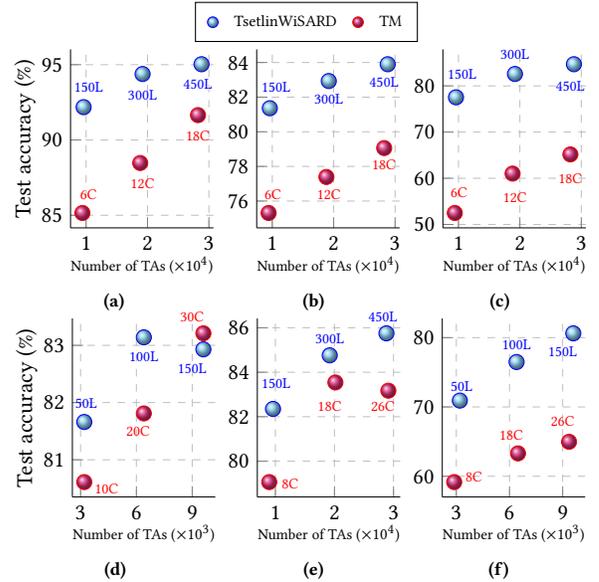

\vspace{-0.5cm}
\section{On-Chip Training Results} \label{sec:on-chip}
\begin{table*}[!b]
%\vspace{-0.2cm}
	\centering
	\caption{Comparisons with FPGA-based WiSARD, TM and CNN accelerators. (\textsuperscript{*}Simulated due to resource limits on target platform. \textsuperscript{**}Used only by Zynq SoC PS and interface for transferring training samples, rather than the on-chip training core.)}
	\label{tab:SoA}
\vspace{-0.3cm}
\resizebox{\textwidth}{!}{%
	\begin{tabular}{|c|c|c||c|c|c||c||c|}
	\hline
	Design & \textbf{TsetlinWiSARD-150} & \textbf{TsetlinWiSARD-300}\textsuperscript{*}  & Hashed WNN \cite{ferreira2019feasible} & LogicWiSARD \cite{miranda2022logicwisard} & BTHOWeN \cite{susskind2022weightless} & DTM \cite{mao2025dynamic} & F-CNN \cite{zhao2016f} \\ \hline
	Platform & \multicolumn{2}{c||}{XC7Z020}  & XC7Z020 & XC7Z010 & XC7Z020 & XC7Z020 &  Stratix V \\ \hline
	Algorithm & \multicolumn{5}{c||}{WiSARD} & TM & CNN \\ \hline
	Purpose & \multicolumn{2}{c||}{Training} & Training & Inference & Inference & Training & Training \\ \hline \hline
	LUTs & 33596  & \cellcolor{gray!66} & 9286 & 2142 & 15756 & 43497 & 69510 \\ \cline{1-2} \cline{4-8}
	FFs & 25927 & \cellcolor{gray!66} & 4568 & 163 & 3522 & 33256 & 87580 \\ \cline{1-2} \cline{4-8}
	\begin{tabular}[c]{@{}c@{}} Other \\ resources \end{tabular} & 	\begin{tabular}[c]{@{}c@{}} 2 BRAMs, \\ 1 DSP\textsuperscript{**} \end{tabular} & \cellcolor{gray!66} N/A & \begin{tabular}[c]{@{}c@{}} 129 BRAMs, \\ 5 DSP, ARM core \end{tabular} & 0 & 0 & \begin{tabular}[c]{@{}c@{}} 138 BRAMs, \\ 6 DSP \end{tabular} & \begin{tabular}[c]{@{}c@{}} 505 BRAMs, \\ 23 DSP \end{tabular} \\ \cline{1-2} \cline{4-8} \cline{1-2} \cline{4-8}
	Latency & \begin{tabular}[c]{@{}c@{}} 6.4 $\mu$s (training \\ \& inference) \end{tabular} & \cellcolor{gray!66} & \begin{tabular}[c]{@{}c@{}} 8.3 ms (training), \\ 9.2 ms (inference) \end{tabular} & 560 ns & 250 ns & \begin{tabular}[c]{@{}c@{}} 95.1 $\mu$s (training), \\ 44.7 $\mu$s (inference)  \end{tabular} &  52.6 ms (training) \\ \cline{1-2} \cline{4-8} \cline{1-2} \cline{4-8}
	Total power (W) & 1.56 & \cellcolor{gray!66} & 0.53 & 0.12 & 0.30 & 4.36 & 27.3 \\  \hline \hline
    	\begin{tabular}[c]{@{}c@{}} Test \\ Accuracy (\%) \end{tabular} & \begin{tabular}[c]{@{}c@{}} 88.02 (MNIST), \\ 79.41 (FMNIST), \\ 72.06 (KMNIST) \end{tabular} & \begin{tabular}[c]{@{}c@{}}  92.1 (MNIST), \\  81.87 (FMNIST), \\  79.97 (KMNIST) \end{tabular} & 90.73 (MNIST) & 91.0 (MNIST) & 93.4 (MNIST) & \begin{tabular}[c]{@{}c@{}} 95.10 (MNIST), \\ 80.6 (FMNIST), \\ 72.5 (KMNIST) \end{tabular} & \cellcolor{gray!66} N/A \\
 	\hline
	\end{tabular}
}
\end{table*}

TsetlinWiSARD is implemented on a Xilinx Zynq XC7Z020 FPGA (PYNQ-Z1), which provides 53,200 LUTs and 106,400 flip-flops (FFs), supporting up to 17,400 LUTRAMs in 28 nm technology. These resources allow deployment of the largest TsetlinWiSARD configuration used in our experiments—featuring 32 TA states, 150 6-input LUTs per discriminator, and 10 discriminators (TsetlinWiSARD-150)—for classification tasks such as MNIST, KMNIST, and FMNIST.
These widely used benchmark datasets enable direct comparison with other ML accelerators. 
A larger model with 300 LUTs per discriminator (TsetlinWiSARD-300) is also constructed and simulated for accuracy comparison; however, its hardware metrics are not reported due to resource limitations preventing practical implementation on the target FPGA.

A Zynq SoC Processing System (PS) IP is implemented to serve as the control unit for data interface. It streams training samples to the TsetlinWiSARD via the AXI bus at runtime.

\begin{figure}[!htb]
\vspace{-0.4cm}
	\centering
	\resizebox{\linewidth}{!}{%
		\begin{tikzpicture}
		[
		pie chart,
		slice type={le}{mypink!50!white},
		slice type={lfsr}{viola!70!white},
		slice type={pop}{giallo},
		slice type={ctrl}{babyblue},
		pie values/.style={font={\normalsize}},
		scale=2
		]		

% LUT
		% le: (17+16*149)*10 = 24010
		% pop: 176*10+103 = 1863
		% lfsr: 26752-24010-1863 = 879
		% ctrl: 33596-26752 = 6844
		\pie[]{33,596 LUTs}{71.5/le,2.6/lfsr,5.5/pop,20.4/ctrl};
		
		% le: 7500 LUTs as memory
		\draw[-stealth, red] (0,1.05) arc[start angle=90, delta angle=80.4, radius=1.05cm];
		% ctrl: 762 LUTs as memory
		\draw[-stealth, red] (0,1.05) arc[start angle=90, delta angle=-8.2, radius=1.05cm];

\draw[
draw=none, % Don't draw arc again
postaction={
	decorate,
	decoration={
		text along path,
		text={|\small\color{red}|24.6 percent as memory},
		text align=center,
		raise=1pt
	}
}
] 
(-1.05,0.25) arc[start angle=170.4, delta angle=-90.4, radius=1cm];
\draw[densely dotted] (0,0) -- ++(81.9:1.1cm);
\draw[densely dotted] (0,0) -- ++(170.4:1.1cm);

\draw[stealth-stealth, blue] (0.15,1.04) arc[start angle=81.9, delta angle=-271.4, radius=1.05cm];
\draw[
draw=none, % Don't draw arc again
postaction={
	decorate,
	decoration={
		text along path,
		text={|\small\color{blue}|75.4 percent as logic},
		text align=center,
		raise=1pt
	}
}
] 
(0.2,1.05) arc[start angle=81.9, delta angle=-80.4, radius=1cm];

% FF
% le: 12*150*10 = 18000
% pop: 19620-18000-1600 = 20
% lfsr: 32*5*10 = 1600
% ctrl: 25927-18000-20-1600 = 6307
\pieN[xshift=2.5cm]{25,927 FFs}{69.4/le,6.2/lfsr,0.1/pop,24.3/ctrl};

% Power
% le: 0.159-0.01-0.025 = 0.124
% pop: 0.001*10 = 0.01
% lfsr: 0.0005*5*10 = 0.025
% ctrl: 1.56-0.124-0.01-0.025 = 1.401
\pieN[xshift=5cm, values of lfsr/.style={pos=0.8, xshift=-0.25cm}, values of pop/.style={pos=1.05}]{1.56 W dynamic power}{7.9/le,0.7/pop,1.6/lfsr,89.8/ctrl};

\legend[shift={(-0.6cm,-1.15cm)}]{{\large{TAs}}/le}
\legend[shift={(0.4cm,-1.15cm)}]{{\large{LFSR}}/lfsr}
\legend[shift={(1.6cm,-1.15cm)}]{{\large{Popcount \& argmax}}/pop}
\legend[shift={(4.05cm,-1.15cm)}]{{\large{Training sample transfer}}/ctrl}
		
		\end{tikzpicture}
	}
\vspace{-0.6cm}
	\caption{TsetlinWiSARD resource and power usage}
\label{fig:breakdown}
\vspace{-0.3cm}
\end{figure}
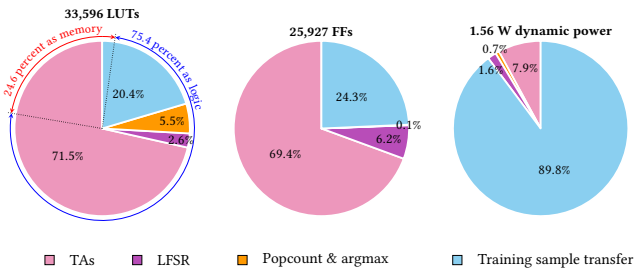

Figure~\ref{fig:breakdown} presents the resource and dynamic power breakdown of the full TsetlinWiSARD-150 implementation. Most LUTs and FFs are consumed by the TAs, enabling a large number of learning elements to maximize classification accuracy. Interestingly, the TsetlinWiSARD logic accounts for only around 10\% of the total dynamic power. This low power footprint is primarily due to minimal data movement—TA states are stored in on-chip LUTRAMs—and the fact that most LUTs operate as static combinational logic, performing simple input-output mappings with limited switching activity.

We present the implementation results of TsetlinWiSARD-150 in Table~\ref{tab:SoA}, with test accuracy evaluated in 100 epochs of on-chip training. Compared to the off-chip training results in Figure~\ref{fig:results_WNNs}(a), the on-chip accuracy is lower. This reduction is primarily due to the pseudo-random behavior of LFSRs: although each stage theoretically yields a 0.5 feedback probability, short-period imbalances can occur, causing certain TAs to receive feedback more frequently and introducing unintended training bias. Despite this, the on-chip accuracy still surpasses that of both standard WiSARD and B-bleaching training methods. The latency reported in Table~\ref{tab:SoA} include both data transfer and training/inference time per sample.

We compare TsetlinWiSARD with other ML accelerators:

1) Compared to the other WiSARD training accelerator, hashed WNNs, TsetlinWiSARD trades LUTs and FFs for reduced usage of BRAMs and DSPs. In hashed WNNs, programmable logic is primarily used to store discriminators, while training and inference are executed by ARM processor, resulting in significantly higher latency. In contrast, TsetlinWiSARD leverages fully parallelized LUTRAMs, enabling over 1000$\times$ speedup in latency. TsetlinWiSARD is the first and only processor-free training accelerator.

2) Compared to existing WiSARD accelerators, TsetlinWiSARD-150 shows slightly lower accuracy, particularly on MNIST. However, as discussed in Section~\ref{sec:param}, accuracy is strongly influenced by architectural factors such as the number of LUTs and the number of inputs per LUT. The higher accuracy achieved by inference-only accelerators like LogicWiSARD and BTHOWeN is largely due to the use of large-scale WiSARD models that are extensively trained and compressed prior to FPGA deployment. Notely, TsetlinWiSARD-300 achieves accuracy on par with the two aforementioned WiSARD implementations.
With access to greater hardware resources, we anticipate the accuracy of TsetlinWiSARD can be further improved.

3) Compared to TM and CNN accelerators, the simplicity of TsetlinWiSARD architecture and its compact hardware implementation enable it to outperform both DTM and F-CNN across all hardware metrics—achieving at least 22.8\% fewer LUTs, 22.0\% fewer FFs, 93.3\% lower training latency, and 64.2\% lower power consumption.
\vspace{-0.3cm}
\section{Conclusion} \label{sec:con}
%\todo{Use past tense or present participle...}
In this work, we proposed TsetlinWiSARD, the first TA-based training method for WiSARD WNNs. Leveraging the TAs' ability to learn in uncertain stochastic environments, we designed a reinforcement-driven training process tailored for WiSARD, achieving higher accuracy than existing WiSARD training methods and outperforming the TA-based TM under comparable resource constraints.

TAs are inherently hardware-friendly due to their simple, discrete state transitions, minimal logic, and highly parallel operations.
Given WiSARD’s structural compatibility with FPGA architecture, both fundamentally constructed using LUTs, TsetlinWiSARD is exceptionally well-suited for efficient on-chip training. We presented a compact FPGA architecture featuring a TA team design and bitwise LFSR, and deployed it on a Xilinx XC7Z020 FPGA. TsetlinWiSARD achieved over 1000$\times$ faster training than existing WiSARD training accelerator and significantly outperformed TM and CNN-based accelerators across all major hardware metrics.
%\todo{remove all one- or two-word lines.}

While WNNs trail recent ML algorithms in accuracy, the ability of TsetlinWiSARD to enable real-time, low-cost, and resource-efficient on-chip learning makes it a compelling solution for edge intelligence and embedded applications.
%\todo{Table 1 needs to go up in page 6. Page 7 for references only.}

\bibliographystyle{ACM-Reference-Format}
\bibliography{sample-base.bib}

\end{document}